\documentclass[10pt, conference]{IEEEtran}
\IEEEoverridecommandlockouts

\usepackage{cite}

\ifCLASSINFOpdf
  \usepackage[pdftex]{graphicx}
  \graphicspath{{./fig/}}
  \DeclareGraphicsExtensions{.pdf,.jpeg,.png}
\else
\fi
\usepackage[caption=false,font=footnotesize]{subfig}

\usepackage{xspace}

\usepackage[utf8]{inputenc}

\usepackage{amssymb}
\usepackage{amsmath, amsthm}
\usepackage{algorithm}
\usepackage{algpseudocode}
\usepackage[hidelinks]{hyperref}

\newcommand*{\super}[1]		{\textsuperscript{#1}}		

\newcommand*{\trackA} 		{\textsc{Track A}\xspace}
\newcommand*{\trackB} 		{\textsc{Track B}\xspace}

\DeclareMathOperator*{\argmin}{arg\,min}

\begin{document}
%
\title{READ-BAD: A New Dataset and Evaluation Scheme for Baseline Detection in Archival Documents}


\author{
\IEEEauthorblockN{Tobias Grüning, Roger Labahn}
\IEEEauthorblockA{Computational Intelligence Technology Lab\\
University of Rostock\\
18057 Rostock, Germany\\
\{tobias.gruening, roger.labahn\}@uni-rostock.de}
\and
\IEEEauthorblockN{Markus Diem, Florian Kleber and Stefan Fiel}
\IEEEauthorblockA{Computer Vision Lab\\
TU Wien\\
1040 Vienna, Austria\\
\{diem,kleber,fiel\}@cvl.tuwien.ac.at}
}


\maketitle

\begin{abstract}
Text line detection is crucial for any application associated with Automatic Text Recognition or Keyword Spotting. Modern algorithms perform good on well-established datasets since they either comprise clean data or simple/homogeneous page layouts. We have collected and annotated $2036$ archival document images from different locations and time periods. The dataset contains varying page layouts and degradations that challenge text line segmentation methods.
Well established text line segmentation evaluation schemes such as the \emph{Detection Rate} or \emph{Recognition Accuracy} demand for binarized data that is annotated on a pixel level. Producing ground truth by these means is laborious and not needed to determine a method's quality. In this paper we propose a new evaluation scheme that is based on baselines. The proposed scheme has no need for binarization and it can handle skewed as well as rotated text lines.
The ICDAR 2017 Competition on Baseline Detection and the ICDAR 2017 Competition on Layout Analysis for Challenging Medieval Manuscripts used this evaluation scheme.
Finally, we present results achieved by a recently published text line detection algorithm.
%
%

\end{abstract}

\IEEEpeerreviewmaketitle


\section{Introduction}
Layout analysis (LA) is considered an open research topic especially for historical collections
and is a major pre-processing step for e.g. Keyword Spotting (KWS) or Handwritten Text Recognition (HTR).
In the last years several competitions were organized to evaluate the performance of layout analysis algorithms: Some focusing
purely on LA \cite{Gatos2011,Antonacopoulos2009,Gatos2010,Antonacopoulos2011,Stamatopoulos2013,Murdock2015}, some requiring a good LA as pre-processing step to achieve competitive
results \cite{Antonacopoulos2013,Antonacopoulos2015,Puigcerver2015}. The ongoing effort in organizing such competitions strongly indicates that there is still
a need for improvement concerning LA.

Even state-of-the-art algorithms have problems if they are faced with degradations related to historical documents \cite{Murdock2015},
e.g. faded-out ink, bleed-through, marginalia, skewed and touching/overlapping text lines.
In contrast, reported results of LA algorithms perform surprisingly well with accuracies far better than $90\%$
\cite{Nicolaou2009,Garz2013,Diem2013,Saabni2014,Chen2015a,boulid2016}. This is basically due to the fact that the well established easily accessible datasets (like
the IAM-HistDB consisting of \textit{Saint Gall Database} \cite{Fischer2011}, \textit{Parzival Database} \cite{Fischer2012} and \textit{Washington Database} \cite{Fischer2012}, as well as the datasets
provided via the competitions \cite{Gatos2011,Gatos2010,Stamatopoulos2013}, the datasets introduced in \cite{Saabni2014} and even newly proposed datasets like the collection of Southeast Asian palm leaf manuscript images \cite{Kesiman2017} are not covering the full range of difficulties present in historical documents.
The datasets contain either modern, well aligned handwritten texts without any serious difficulties for state-of-the-art algorithms at all
or very homogeneous layouts within a dataset, hence it is an ease to adapt algorithms to such datasets.


Since state-of-the-art methods achieve high accuracies on well-established datasets, there is a need for a new, challenging dataset with complex page layouts and a greater variety in terms of script, time range and place of origin.
A huge variety of degradations as well
as different resolutions and orientations should be present. Since the landscape of document analysis has changed over the last years, and machine learning based algorithms get more and more popular not only for
KWS \cite{strauss2016citlab} and HTR \cite{Graves2008} but also for LA \cite{Moysset2015,Chen2016,Pastor-Pellicer2016}, the dataset should consist of hundreds of pages to provide an appropriate amount of training samples.

Besides the characteristics of the images the kind of ground truth (GT) provided is essential. The variety of GT given for different datasets ranges from origin points \cite{Murdock2015}
over polygons surrounding the text lines \cite{Fischer2011,Fischer2012} and ground truth on pixel level \cite{Gatos2011,Gatos2010,Saabni2014} to detailed information about text region entities \cite{Antonacopoulos2011} and reading order \cite{Antonacopoulos2015}.
Since in the most application scenarios LA is mainly a pre-processing step for HTR, it is meaningful to provide goal-oriented GT. Modern HTR systems require text lines as input \cite{Graves2008,strauss2016citlab},
that is why we will restrict ourselves to the text line detection scenario and ignore issues like entity classification and reading order. Nevertheless in complex layout scenarios (e.g. tables, multi-column texts, present marginalia), it is mandatory to detect the page
layout to achieve correct text line segmentation results. Ignoring the page layout typically leads to an undersegmentation of text lines, see Sec.~\ref{sec:bs}. Therefore, the text line segmentation scenario somehow comprises the page segmentation scenario as a required intermediate processing step.

To characterize the text lines using solely origin points is in our opinion not sufficient since they don't cover the characteristics, e.g. skew, orientation, dimension, ... , of the text lines at all. On the other hand, \cite{7333819} showed that the HTR accuracy is not significantly effected by the polygon surrounding the text lines. Even simple
strategies to construct surrounding polygons given baseline representations lead to satisfying results \cite{7333819}. Therefore, GT based on baseline representations for the text lines is in our opinion a reasonable compromise. Furthermore, annotating baselines is less cumbersome than surrounding polygons and therefore cheaper.

Since the widely-used evaluation schemes rely on surrounding polygons and use area (or foreground pixel) based methods to calculate the accuracy of text line segmentation results, there is a need for an evaluation scheme suitable for baselines.

In this paper, we introduce a new dataset containing $2036$ pages of historical documents with annotated baselines. Furthermore, we propose a newly developed,
goal-oriented evaluation scheme working with baseline representations of the text lines. This scheme was already used in two layout analysis competitions, namely the ICDAR $2017$ Competition on Baseline Detection (cBAD) and the ICDAR $2017$ Competition on Layout Analysis for Challenging Medieval Manuscripts. While we published a report of the cBAD competition alongside with the evaluation scheme 
in~\cite{diem2017}, this paper aims at a thorough introduction of the evaluation scheme. In addition, the collection of the dataset and its sources are described.

The remaining paper is structured as follows, in Section \ref{sec:data} the dataset is described, a meaningful subdivision is explained and some example pages as well as statistics are shown. Section \ref{sec:eval} describes the newly proposed evaluation scheme along with some examples demonstrating the functionality of the scheme. In Section \ref{sec:bs} the results obtained by a recently published
text line detection method are presented. Section \ref{sec:conc} concludes the paper.

\section{Dataset}
\label{sec:data}
The ICDAR $2017$ Competition on Baseline Detection (cBAD) dataset \cite{diem_markus_2017_257972} is composed of $2036$ document page images that were collected from $9$ different archives.
\subsection{Baseline Definition}
A baseline is defined in the typographical sense as the virtual line where most characters rest upon and descenders extend below.
Text lines are annotated by one single baseline.
Hence, non-textual symbols are not annotated. Non-textual symbols include: decoration lines, dotted lines, images, noise/stains, initials, bleed-through text.
A baseline is split if
\begin{itemize}
\item it spans different columns.
\item it spans different document pages.
\item it connects marginalia and the body text.
\end{itemize}
If a text line is clearly not part of a table (column) system, a single baseline is annotated even crossing column borders.


\subsection{The cBAD Dataset}
About $2000$ document images from each of $9$ different European archives were collected. These documents were written between $1470$ and $1930$.
We sampled $250$ images from each archival collection using a freely available python script\footnote{https://github.com/TUWien/Benchmarking}. This results in a set of $2500$ document images.
A more detailed description of the $9$ different document collections is given below.

\emph{Archive Bistum Passau (ABP):} collection contains $16,000$ images photographed at $300$ dpi. The documents include parish registers of baptisms, marriages, and funerals.


\emph{Bohisto - Bozen State Archive:} $77,000$ page images of council minutes written between $1470$ and $1804$.

\emph{Venice Time Machine (EPFL):} about $5000$ pages from indexes of records, records of real property transactions, and daily death registrations written between the $16$\super{th} and $18$\super{th} century.

\emph{Humboldt University Berlin (HUB):} $3600$ student notes of lectures given by Alexander von Humboldt between $1827$ and $1829$.

\emph{National Archive Finland (NAF):} $2186$ page images from account books, a court book, a census book, and a church book that cover a time period from $1774$ until the $1930$s.

\emph{Marburg State Archive:} $36,000$ page images from the Grimm collection comprising letters, postcards, and greeting cards.

\emph{University College London (UCL):} the Bentham papers include $55,000 pages$. Most pages were written by the British philosopher Jeremy Bentham between $1760$ and $1832$.

\emph{Brabant Archive (BHIC):} composed of various types of tables containing census information.

\emph{University Bibliography Basel (unibas):} e-manuscripta\footnote{\url{http://www.e-manuscripta.ch/}}.


\begin{figure}[ht]
	\centering
	\includegraphics[width=0.48\columnwidth]{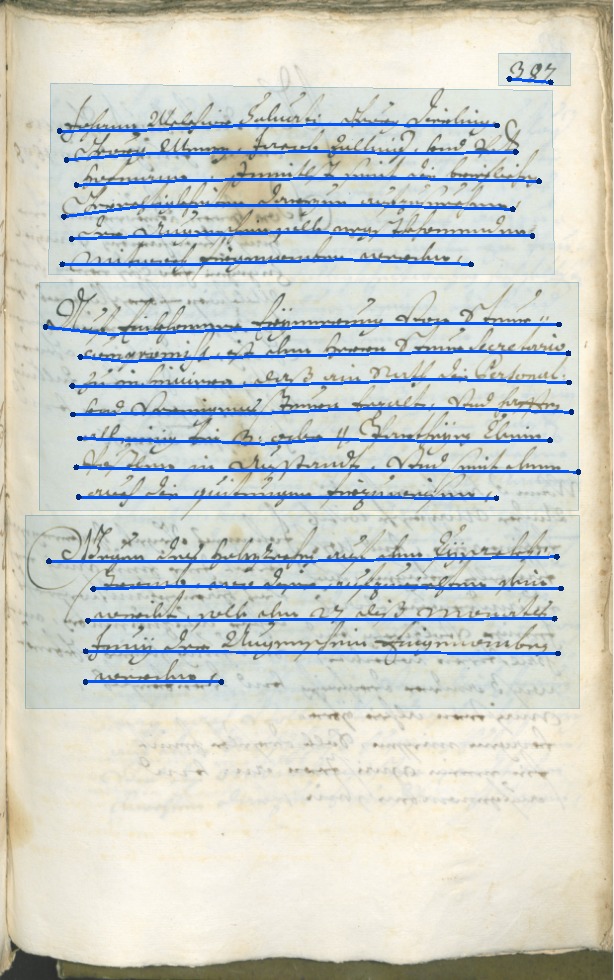}
	\includegraphics[width=0.5\columnwidth]{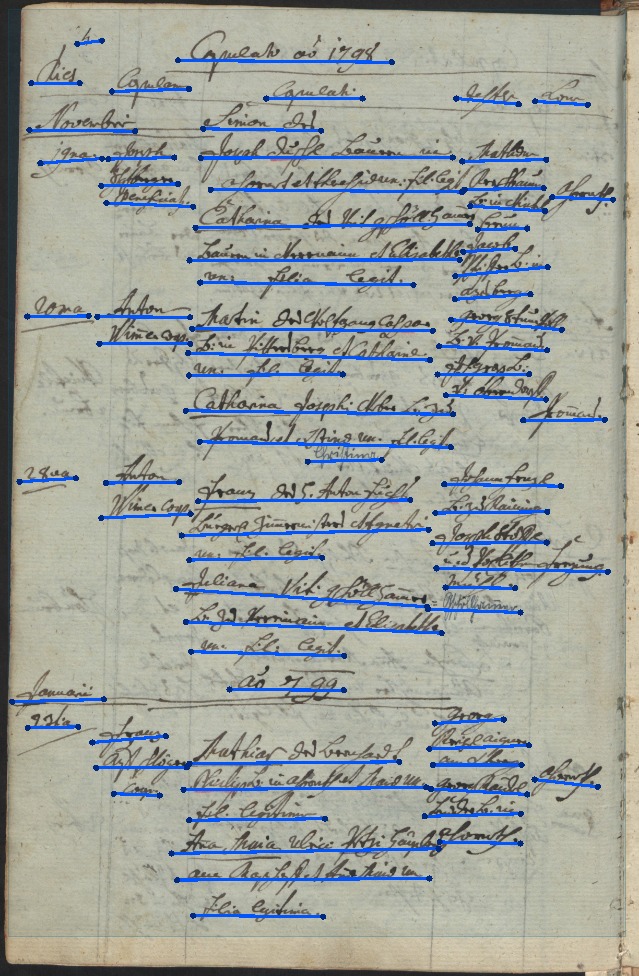}
	\caption{Two example images of \trackA Simple Documents (left) and \trackB Complex Documents (right) with annotated baselines and text regions.}
	\label{fig:images}
\end{figure}

\subsection{Data Annotation}
After removing images due to quality as well as content issues the number reduced from $2250$ to $2118$. For these images the text regions as well as baselines were annotated by DigiTexx.
The well-known PAGE XML\footnote{\url{http://www.primaresearch.org/tools}} scheme is used for storing text region and baseline information.
A final review process by two independent operators reduced the total number to $2036$ images. All in all $132,124$ annotated baseline are available.

This annotated dataset is split into two subsets: \emph{Simple Documents} and \emph{Complex Documents}. The first includes only pages with simple page layouts and annotated text regions.
Hence, this could be used for a track to evaluate the text line segmentation only, thus neglecting issues that arise from the page layout.
The second subset \emph{Complex Documents} includes full page tables, multi column text and rotated text lines. The challenge is not only to robustly detect baselines but also to split baselines correctly
with respect to the page layout.

Both subsets are split into a training and a test set. For training $30$ images are taken from each collection resulting in $216$ training images for \emph{Simple Documents} and $270$ images for \emph{Complex Documents}.
The data along with the GT is publicly available\cite{diem_markus_2017_257972}. Two example images are shown in Fig.~\ref{fig:images}.

\section{Evaluation Scheme}
\begin{figure*}[t]
\centering
\subfloat[Depicted are the sets $\mathcal{G}=\{g_1,g_2,g_3,g_4\}$ of four GT baselines (blue) and $\mathcal{H}=\{h_1,h_2,h_3,h_4\}$ of four HY baseline (red).]{\includegraphics[width=0.95\textwidth]{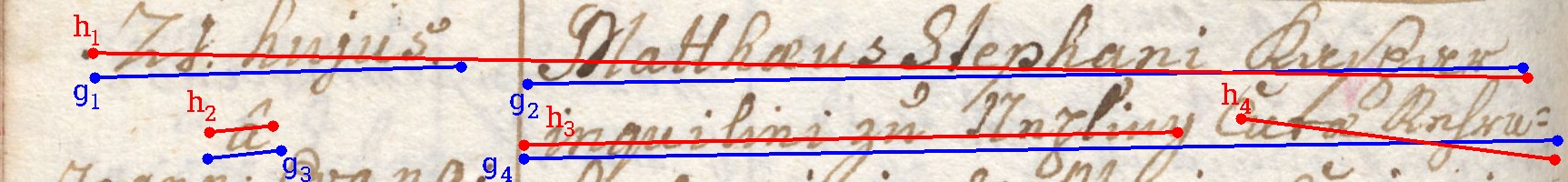}
\label{fig:in}}
\hfil
\subfloat[The same baselines as shown in Fig.~\ref{fig:in} but represented by normalized polygonal chains (for better clarity only every $25$th vertex is shown).]{\includegraphics[width=0.95\textwidth]{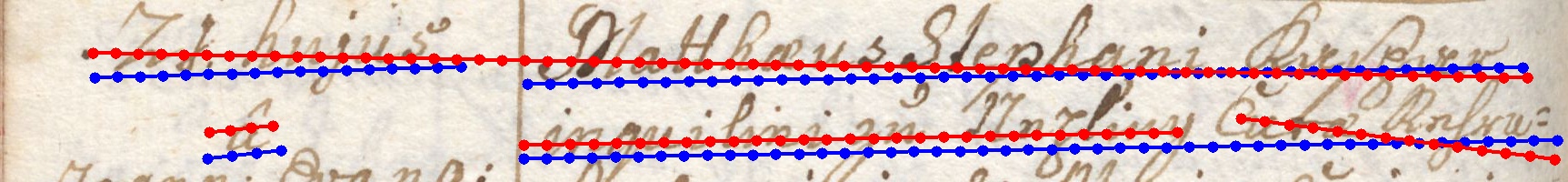}
\label{fig:norm}}
\hfil
\subfloat[The vertices, which are taken into account to calculate the minimum distance of GT line $2$ to the other GT lines, are displayed as green points. Green lines are the orthogonal (to GT line $2$) distances of the green vertices.]{\includegraphics[width=0.95\textwidth]{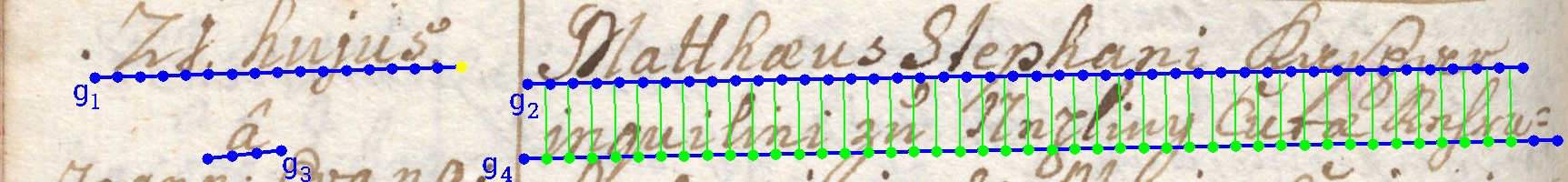}
\label{fig:dist}}
\hfil
\subfloat[Shown in light blue are the tolerance areas for the different GT baselines, for all four baselines the estimated tolerance value is roughly $20$.]{\includegraphics[width=0.95\textwidth]{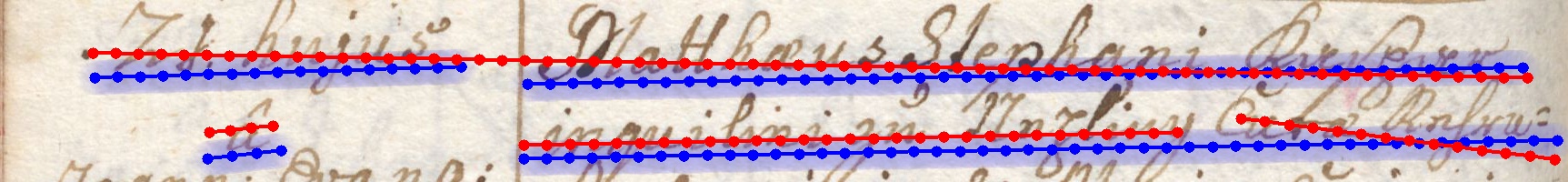}
\label{fig:tol}}
\caption{Depicted is a snippet of an example document sampled from \cite{diem_markus_2017_257972} and intermediate steps of the evaluation scheme.}
\end{figure*}

\label{sec:eval}
Since baseline detection is the first step in the information retrieval pipeline of an classical workflow, there are special requirements regarding the evaluation scheme:
\begin{itemize}
 \item The evaluation scheme should indicate how reliable the text is detected -- ignoring layout issues. The value reflecting this is called \emph{R-value}, since it has similar properties as the well-known recall value.
 \item The evaluation scheme should indicate how reliable the structure of the text lines (layout) of the document is detected. The value reflecting this is called \emph{P-value}, since it has similar properties as the well-known precision value.
 \item The evaluation scheme should be invariant to small differences between ground truth and hypotheses. There is not an unique correct baseline, slightly different baselines potentially lead to the same HTR accuracy.
 \item The evaluation scheme should be able to handle skewed and oriented text lines
 \item The evaluation scheme should not rely on a reading order nor on a binarization

\end{itemize}
To our knowledge there is no evaluation scheme meeting these requirements -- or even any scheme working for baselines. Hence, we propose a newly developed scheme to evaluate the performance of baseline detection algorithms.
The proposed algorithm is implemented in Java and available as a standalone command line tool. It is licensed under LGPLv3 and publicly available\footnote{https://github.com/Transkribus/TranskribusBaseLineEvaluationScheme}.

\subsection{Single Page Evaluation}
In the following the calculation of R and P for a single page is explained.
Let $\mathcal{P}$ be the set of all polygonal chains (each polygonal chain represents a baseline and contains a finite number of ordered vertices, which are characterized by two coordinates).  $\mathcal{G} = \left\{\boldsymbol{g}_1,...,\boldsymbol{g}_M\right\}\subset \mathcal{P}$ is the set of given (GT) polygonal chains representing the baselines for a single page and
$\mathcal{H} = \left\{\boldsymbol{h}_1,...,\boldsymbol{h}_K\right\}\subset \mathcal{P}$ is the set of hypothesis (HY) polygonal chains calculated by a baseline detection algorithm for the same page, Fig.~\ref{fig:in}.
The calculation of R and P for the two sets $\mathcal{G}$ and $\mathcal{H}$ follows:

\subsubsection{Polygonal Chain Normalization}
In a first step each chain is normalized, so that two adjacent vertices are in the $8$-neighborhood of each other (have a distance $\leq \sqrt{2}$), Fig.~\ref{fig:norm}. The resulting sets of normalized chains are $\widetilde{\mathcal{G}}$ and $\widetilde{\mathcal{P}}$. For better readability we omit the tilde. In the following $\mathcal{G}$ and $\mathcal{P}$ are the sets
of normalized polygonal chains.

\subsubsection{Tolerance Value Calculation}
In a second step for each chain $\boldsymbol{g}\in\mathcal{G}$ a \textit{tolerance value} $t_{\boldsymbol{g}}$ is calculated. As mentioned above,
the evaluation scheme should not penalize HY baselines which are slightly different to the GT baselines. Hence, some kind of tolerance is necessary.
Page (and text line) dependent tolerance values are calculated, because within a collection various resolutions and layout scenarios could be present.
A single pre-defined tolerance value can hardly cover all these scenarios in a satisfying fashion.
Since the $y$-coordinates of the vertices are typically ``wrongly'' oriented in computer vision scenarios, they have to be negated for the following procedure.
To calculate $t_{\boldsymbol{g}}$, the orientation $\alpha_{\boldsymbol{g}}\in [0,\pi)$ of $\boldsymbol{g}$ is estimated using linear regression.
$\boldsymbol{o}(\alpha_{\boldsymbol{g}})=(\cos(\alpha_{\boldsymbol{g}}),\sin(\alpha_{\boldsymbol{g}}))^T$ is the vector of length $1$ of orientation $\alpha_{\boldsymbol{g}}$.
Given the set $\mathcal{V}$ of all vertices
of the chains in $\mathcal{G}\setminus \boldsymbol{g}$, the subset $\mathcal{V}_{\boldsymbol{g}}\subset\mathcal{V}$ is calculated such that for any $\boldsymbol{v}\in\mathcal{V}_{\boldsymbol{g}}$ there are at least
two vertices $\boldsymbol{v}_1,\ \boldsymbol{v}_2\in\boldsymbol{g}$ satisfying
\begin{align}
\label{eq:ind}
 (\boldsymbol{v}-\boldsymbol{v}_1)^T\boldsymbol{o}(\alpha_{\boldsymbol{g}})\cdot (\boldsymbol{v}-\boldsymbol{v}_2)^T\boldsymbol{o}(\alpha_{\boldsymbol{g}}) \leq 0.
\end{align}
Condition (\ref{eq:ind}) means that the projections of $(\boldsymbol{v}-\boldsymbol{v}_1)$ and $(\boldsymbol{v}-\boldsymbol{v}_2)$ into the direction of $\boldsymbol{o}(\alpha_{\boldsymbol{g}})$ have different algebraic signs (or have length zero).
In Fig.~\ref{fig:dist} the set $\mathcal{V}_{\boldsymbol{g}_2}$ of vertices for GT baseline $2$ is shown (green points). For each $\boldsymbol{v}\in\mathcal{V}_{\boldsymbol{g}}$ one vertex $\boldsymbol{v}_{m}(\boldsymbol{v})\in\boldsymbol{g}$ is determined
for which the projection of $(\boldsymbol{v}-\boldsymbol{v}_{m}(\boldsymbol{v}))$ into the direction of $\boldsymbol{o}(\alpha_{\boldsymbol{g}})$ has minimal length
\begin{align*}
 \boldsymbol{v}_{m}(\boldsymbol{v})=\argmin_{\boldsymbol{v}_{\boldsymbol{g}}\in \boldsymbol{g}} \left|(\boldsymbol{v}-\boldsymbol{v}_{\boldsymbol{g}})^T\boldsymbol{o}(\alpha_{\boldsymbol{g}}) \right|.
\end{align*}
The minimum distance of $\boldsymbol{g}$ to another chain is calculated by
\begin{align*}
d_{\boldsymbol{g}}=\min_{\boldsymbol{v}\in\mathcal{V}_{\boldsymbol{g}}}\left|(\boldsymbol{v}-\boldsymbol{v}_{m}(\boldsymbol{v}))_x \boldsymbol{o}(\alpha_{\boldsymbol{g}})_y - (\boldsymbol{v}-\boldsymbol{v}_{m}(\boldsymbol{v}))_y\boldsymbol{o}(\alpha_{\boldsymbol{g}})_x \right|.
\end{align*}
Subscripts $\boldsymbol{v}_x$ and $\boldsymbol{v}_y$ are the $x$- and $y$-coordinate of vector $\boldsymbol{v}$.
$d_{\boldsymbol{g}}$ is the minimal length of the projections of all $(\boldsymbol{v}-\boldsymbol{v}_{m}(\boldsymbol{v}))$ into the direction orthogonal to $\boldsymbol{o}(\alpha_{\boldsymbol{g}})$, see Fig.~\ref{fig:dist} (green lines).
For $\mathcal{V}_{\boldsymbol{g}}=\emptyset$ there are no other baselines allowing a meaningful calculation of $d_{\boldsymbol{g}}$, hence its tolerance value is set to some default value ($250$ was chosen).
Condition (\ref{eq:ind}) is essential since $\mathcal{V}_{\boldsymbol{g}}$ is the basis
for the estimation of the minimal distance of $\boldsymbol{g}$ to another chain. For instance the yellow vertex Fig.~\ref{fig:dist} has a significantly shorter orthogonal projection to GT line $2$,
but of course would falsify the statistics. The mean $\overline{d}_{\mathcal{G}}$ of all $d_{\boldsymbol{g}}$ ($\boldsymbol{g}\in\mathcal{G}$) with a value different to the default value is calculated. Finally, the GT baseline dependent tolerance
values are calculated
\begin{align*}
t_{\boldsymbol{g}} = 0.25\cdot\min(d_{\boldsymbol{g}},\overline{d}_{\mathcal{G}}).
\end{align*}
$25 \%$ of the estimated interline distance yields a reasonable compromise between accuracy and flexibility.
$\mathcal{T}=\mathcal{T}(\mathcal{G})$ is the set  containing the resulting tolerance values, in Fig.~\ref{fig:tol} the blue areas show the individual tolerance areas for the different GT baselines.

\subsubsection{Coverage Function}
Employing the (tolerance dependent) $\text{COV}:\mathcal{P}\times\mathcal{P}\times\mathbb{R}\rightarrow\mathbb{R}$ function implemented via Alg.~\ref{alg:cnt}, one can determine a value
representing the fraction of chain $\boldsymbol{p}$ for which there is a vertex of chain $\boldsymbol{q}$ within a certain tolerance area (skew-invariant).
\begin{algorithm}[t]
    \caption{Coverage Function}
    \label{alg:cnt}
    \begin{algorithmic}[1]
        \Procedure{cov}{$\boldsymbol{p},\boldsymbol{q},t$}
        \State $c\gets 0$
        \For{$p=(p_x,p_y)$ vertex of $\boldsymbol{p}$}
        \State $d_{min} \gets \min_{q\in\boldsymbol{q}}(\left\|p-q\right\|_2)$
	  \If{$d_{min} \leq t$}
	   \State $c\gets c+1$
	  \ElsIf{$d_{min} \leq 3 t$}
	  \State $c\gets c+\frac{3  t - d_{min}}{2 t}$
	  \EndIf
        \EndFor
        \State $c \gets \frac{c}{\left| \boldsymbol{p} \right|}$ \Comment{$\left| \boldsymbol{p} \right|$ is the number of vertices of $\boldsymbol{p}$}
         \State \textbf{return} $c$
        \EndProcedure
    \end{algorithmic}
\end{algorithm}
Alg.~\ref{alg:cnt} counts the number of vertices of $\boldsymbol{p}$ for which there is a vertex of $\boldsymbol{q}$ with a distance less than the given tolerance value $t$.
Furthermore a smooth (linear) transition is performed for vertices with a distance between $t$ and $3t$. A vertex with a distance less than $t$ counts $1$, with a distance of $1.5t$ it counts $0.75$,
with a distance of $2t$ it counts $0.5$, ... Finally, a vertex with a distance of $3t$ and more counts $0$. The resulting value is normalized using the number of vertices of $\boldsymbol{p}$.

Let $\text{COV}_S:\mathcal{P}\times\mathfrak{P}(\mathcal{P})\times\mathbb{R}\rightarrow\mathbb{R}$ be the generic extension of $\text{COV}$ to a function accepting sets of polygonal chains as second argument.
The minimum from line $4$ in Alg.~\ref{alg:cnt} is calculated over a set of chains instead of a single chain.
To clarify the functionality of the coverage functions a few exemplary values are shown in Tab.~\ref{tab:cover}. Especially, the function $\text{COV}$ is not commutative in the first two arguments.

\begin{table}[ht]
\renewcommand{\arraystretch}{1.3}
\caption{Example values of the coverage functions applied to normalized polygonal chains shown in Fig.~\ref{fig:norm} with a fixed tolerance value of $20$ (as shown in Fig.~\ref{fig:tol}). $g_i$ means the normalized version of the $i$-th GT baseline. }
\label{tab:cover}
\centering
\begin{tabular}{c||c||c||c||c}
\hline
$\boldsymbol{p}$ & $\boldsymbol{q}$ & $\boldsymbol{r}$ & $\text{COV}(\boldsymbol{p},\boldsymbol{q},20)$ & $\text{COV}_S(\boldsymbol{p},\{\boldsymbol{q},\boldsymbol{r}\},20)$\\
\hline\hline
$h_3$ & $g_4$ & -- & $1.0$ & --\\
$g_4$ & $h_3$ & $h_4$ & $0.65$ & $0.96$\\
$g_3$ & $h_2$ & -- & $0.76$ & --\\
$h_1$ & $g_1$ & $g_2$ & $0.26$ & $0.95$\\
\hline
\end{tabular}
\end{table}

\subsubsection{R and P Calculation}
The tolerance dependent R
value of $\mathcal{G}$ and $\mathcal{H}$ is finally calculated by
\begin{align}
\label{eq:rec}
 \text{R}(\mathcal{G},\mathcal{H},\mathcal{T}) = \frac{\sum_{\boldsymbol{g}\in\mathcal{G}}\text{COV}_S(\boldsymbol{g},\mathcal{H},t_{\boldsymbol{g}})}{\left|\mathcal{G} \right|}.
\end{align}
The R value indicates for what fraction of the GT baselines there are detected HY baselines within a certain tolerance area.
Segmentation (page layout) errors are not penalized at all, because no alignment between GT and HY baselines is enforced.

These segmentation errors are penalized in the P value. Let $\mathcal{M}(\mathcal{G},\mathcal{H}) \subset \mathcal{G}\times\mathcal{H}$ be  an alignment of GT and HY chains
where each element of $\mathcal{G}$ as well as of $\mathcal{H}$ occurs at most once. The tolerance dependent P value of $\mathcal{G}$ and $\mathcal{H}$ is calculated as follows
\begin{align}
\label{eq:prec}
 \text{P}(\mathcal{G},\mathcal{H},\mathcal{T}) = \frac{\sum_{(\boldsymbol{g},\boldsymbol{h})\in\mathcal{M}(\mathcal{G},\mathcal{H})}\text{COV}(\boldsymbol{h},\boldsymbol{g},t_{\boldsymbol{g}})}{\left|\mathcal{H} \right|}.
\end{align}
An alignment ensures that segmentation errors are penalized. E.g. if a text line is split into two equally sized parts, a R value of
$1.0$ is calculated (the two detected chains cover the entire GT chain), but the expected P value is $0.5$ (the GT chain is aligned with exactly one of
the HY chains with a P value of $1$, this is divided by $2$, because there are two HY chains). We want to mention that for both cases (R and P)
short text lines have the same impact as long ones, because in (\ref{eq:rec}) and (\ref{eq:prec}) the line specific R and P values are divided by the number of GT respectively HY lines. This prevents the proposed evaluation scheme from
underestimating the importance of short text lines, which often contain essential information in the context of historical documents, e.g. dates.

\subsubsection{Greedy-based Alignment}
To evaluate (\ref{eq:prec}) an P-optimal alignment is necessary. Therefore a P matrix $\boldsymbol{C}\in \mathbb{R}^{M\times K}$ is calculated with elements $c_{ij}=\text{COV}(\boldsymbol{h}_i, \boldsymbol{g}_j, t_{\boldsymbol{g}_j})$.
Based on this, the alignment is calculated in a greedy manner $\mathcal{M}(\mathcal{G},\mathcal{H}) = \text{ALIGN}(\boldsymbol{C},\mathcal{G},\mathcal{H})$, see Alg.~\ref{alg:match}. A greedy approach was chosen, because there is no reading order available
(no dynamic programming possible) and the greedy solution is in most practical cases the exact solution.
\begin{algorithm}[ht]
    \caption{Alignment Function}
    \label{alg:match}
    \begin{algorithmic}[1]
        \Procedure{align}{$\boldsymbol{C},\mathcal{G},\mathcal{H}$}
        \State $\mathcal{M} \gets \emptyset$
        \State $\boldsymbol{C}' \gets \boldsymbol{C}$
        \While{$\boldsymbol{C}'$ is not empty}
        \State $m \gets$ one of the maximal elements of $\boldsymbol{C}' $
        \If{$m>0$}
	  \State //create a new matching pair
	  \State $\boldsymbol{g} \gets$ element of $\mathcal{G}$ belonging to $m$
	  \State $\boldsymbol{h} \gets$ element of $\mathcal{H}$ belonging to $m$
	  \State $\mathcal{M} \gets \mathcal{M} \cup (\boldsymbol{g},\boldsymbol{h})$
	  \State $\boldsymbol{C}' \gets$ take $\boldsymbol{C}'$ and delete row/col of $m$
	  \Else
	   \State \textbf{return} $\mathcal{M}$
        \EndIf
        \EndWhile
        \State \textbf{return} $\mathcal{M}$
        \EndProcedure
    \end{algorithmic}
\end{algorithm}
\subsubsection{Harmonic Mean (F value)}
Finally, the harmonic mean of R and P, we call it F value,
\begin{align*}
 \text{F} = \frac{2\cdot \text{R}\cdot \text{P}}{\text{R} + \text{P}}
\end{align*}
is calculated.

\subsection{Multi Page Evaluation}
Since the dataset is very heterogeneous, each page is evaluated on its own. The average is calculated for this page-wise results. This prevents an overbalance of pages with dozens of baselines (like pages containing a table)
and yields results representing the robustness of the evaluated algorithms over various scenarios.

\subsection{Examples}
Results for different subsets of the GT and HY baselines of Fig.~\ref{fig:in} are shown in Tab.~\ref{tab:ex} and explained in the following.
\begin{table}[ht]
\renewcommand{\arraystretch}{1.3}
\caption{Example values for R, P and $F_1$ for different subsets of the GT and HY baselines shown if Fig.~\ref{fig:in}, for all evaluations the tolerance parameter was fixed to $20$.}
\label{tab:ex}
\centering
\begin{tabular}{c||c||c||c||c||c}
\hline
Ex. & $\mathcal{G}$ & $\mathcal{H}$ & R & P & $F_1$\\
\hline\hline
1& $\{g_1,g_2,g_3,g_4\}$ & $\{h_1,h_2,h_3,h_4\}$ & $0.91$ & $0.61$ & $0.73$\\
2& $\{g_2,g_3,g_4\}$ & $\{h_1,h_2,h_3,h_4\}$ & $0.9$ & $0.61$ & $0.73$\\
3& $\{g_1,g_3,g_4\}$ & $\{h_1,h_2,h_3,h_4\}$ & $0.89$ & $0.51$ & $0.65$\\
4& $\{g_1,g_2,g_3,g_4\}$ & $\{h_2,h_3\}$ & $0.35$ & $0.88$ & $0.5$\\
5& $\{g_1,g_2,g_3,g_4\}$ & $\{h_2,h_3,h_4\}$ & $0.43$ & $0.6$ & $0.5$\\
\hline
\end{tabular}
\end{table}
The small difference between Ex. $1$ and Ex. $2$ is due to the fact, that in both cases $h_1$ is aligned to $g_2$ for the P calculation.
Hence, there is no effect on P if $g_1$ is removed. R is nearly the same, because $g_1$ and $g_2$ are both completely covered by $h_1$.
By removing $g_2$ instead of $g_1$ (Ex. $3$), $h_1$ is now aligned to $g_1$ yielding a lower P value, because $g_2$ covers much more of $h_1$ than $g_1$.
In Ex. $4$ one gets a high P value, because the remaining HY baselines are very well covered by the GT baselines. By adding $h_4$ (Ex. $5$) we of course increase R, but decrease P.
This is due to the fact that $h_3$ is aligned to $g_4$ (as in Ex. $4$) and $h_4$ is not aligned at all and gets a P value of $0$.

\section{Baseline System}
\label{sec:bs}
In this section we present the results obtained by applying the text line detection algorithm presented in \cite{gr2017}. This approach relies on the clustering of so-called superpixels (SPs). These SPs were calculated utilizing the classical FAST
algorithm. The algorithm does not rely on any training process. Hence, the training subset was ignored and the proposed algorithm was just applied for the test subset (without any parameter tuning). The results obtained are depicted in Tab.~\ref{tab:results}.
\begin{table}[H]
\renewcommand{\arraystretch}{1.3}
\caption{Results obtained for the simple and complex test-sets by the method proposed in \cite{gr2017}.}
\label{tab:results}
\centering
\begin{tabular}{c||c|c|c|c|c}
\hline
Track & GT lines & HY lines & R & P & $F_1$\\
\hline\hline
Simple & $14735$ & $15836$ & $0.941$ & $0.884$ & $0.912$\\
Complex & $88962$ & $50166$ & $0.67$ & $0.795$ & $0.728$\\
\hline
\end{tabular}
\end{table}
As mentioned in \cite{gr2017} the method struggles if faced with complex layouts. The method suffers from undersegmentation problems and results in a bad accuracy for the complex track compared to the simple track.

\section{Conclusion}
\label{sec:conc}
A new dataset consisting of $2036$ pages of archival documents with $132,124$ annotated baselines was introduced. A wide span of different times as well as locations is covered. The dataset contains documents with various degradations and complex layouts.
Along with the dataset a goal-oriented evaluation scheme based on baseline representations is introduced. Finally, the results obtained by a baseline system are shown.
This work provides new challenges as well as a solid basis for competitive evaluations
for the document layout community.

\section*{Acknowledgment}
This work was partially funded by the European Union's Horizon $2020$ research and innovation programme under grant agreement No $674943$ (READ -- Recognition and Enrichment of Archival Documents).

\bibliographystyle{IEEEtran}
%




%

\bibliography{IEEEabrv,lit}

\end{document}